\documentclass[letterpaper, 10 pt, conference]{ieeeconf-modified}

\IEEEoverridecommandlockouts

\usepackage[numbers]{natbib}
\usepackage{graphicx}
\graphicspath{ {./figures} }
\usepackage{siunitx}
\usepackage{algorithm,algpseudocode}
\usepackage{amsmath}
\usepackage{amssymb} 
\usepackage{textcomp}
\usepackage{hyperref}
\usepackage[capitalize]{cleveref}
\usepackage{makecell}
\usepackage{booktabs}
\usepackage[dvipsnames]{xcolor}
\usepackage{tablefootnote}

\usepackage{fancyhdr}

\fancypagestyle{FirstPage}{
\lfoot{\fontsize{7}{7}\selectfont \begin{center}\copyright2023 IEEE. Personal
use of this material is permitted. \\Permission from IEEE must be
obtained for all other uses, in any current or future media,
including reprinting/republishing this material for advertising or
promotional purposes, creating new collective works, for resale or
redistribution to servers or lists, or reuse of any copyrighted
component of this work in other works.\end{center}}
}

\crefname{equation}{}{}
\newcommand{\gitlink}{https://dlr-alr.github.io/dlr-tactile-manipulation}
\newcommand{\tentry}[2]{$#1$ \tiny{$\pm #2$}}

\title{\LARGE \bf
Estimator-Coupled Reinforcement Learning \\
for Robust Purely Tactile In-Hand Manipulation
}
\author{Lennart Röstel, Johannes Pitz, Leon Sievers and Berthold Bäuml%
\thanks{The authors are with the DLR Institute of Robotics and Mechatronics, Technical University of Munich and  Deggendorf Institute of Technology. \newline
Contact: \tt\footnotesize{lennart.roestel@dlr.de}
}
}

\begin{document}

\maketitle
\thispagestyle{FirstPage}
\pagestyle{empty}

\begin{abstract}
This paper identifies and addresses the problems with naively combining (reinforcement) learning-based controllers and state estimators for robotic in-hand manipulation. Specifically, we tackle the challenging task of purely tactile, goal-conditioned, dextrous in-hand reorientation with the hand pointing downwards.
Due to the limited sensing available, many control strategies that are feasible in simulation when having full knowledge of the object's state do not allow for accurate state estimation. Hence, separately training the controller and the estimator and combining the two at test time leads to poor performance. 
We solve this problem by coupling the control policy to the state estimator already during training in simulation.
This approach leads to more robust state estimation and overall higher performance on the task while maintaining an interpretability advantage over end-to-end policy learning. 
With our GPU-accelerated implementation, learning from scratch takes a median training time of only 6.5 hours on a single, low-cost GPU.
In simulation experiments with the DLR-Hand~II and for four significantly different object shapes, we provide an in-depth analysis of the performance of our approach. 
We demonstrate the successful sim2real transfer by rotating the four objects to all 24 orientations in the $\pi/2$ discretization of SO(3), which has never been achieved for such a diverse set of shapes. 
Finally, our method can reorient a cube consecutively to nine goals (median), which was beyond the reach of previous methods in this challenging setting.
 \\
Website: \href{\gitlink}{\scriptsize\texttt{\gitlink}}
\end{abstract}

\section{Introduction}
\label{sec:intro}
State estimation and control are at the core of many problems in robotics. 
Although the performance of one component is highly dependent on the quality of the other, controllers and state estimators are usually derived separately and often combined only at the end of the development process. 
Consequently, the predicted control inputs are computed under the assumption of accurate state input.

\begin{figure}
  \centering
  \includegraphics[width=.8\linewidth]{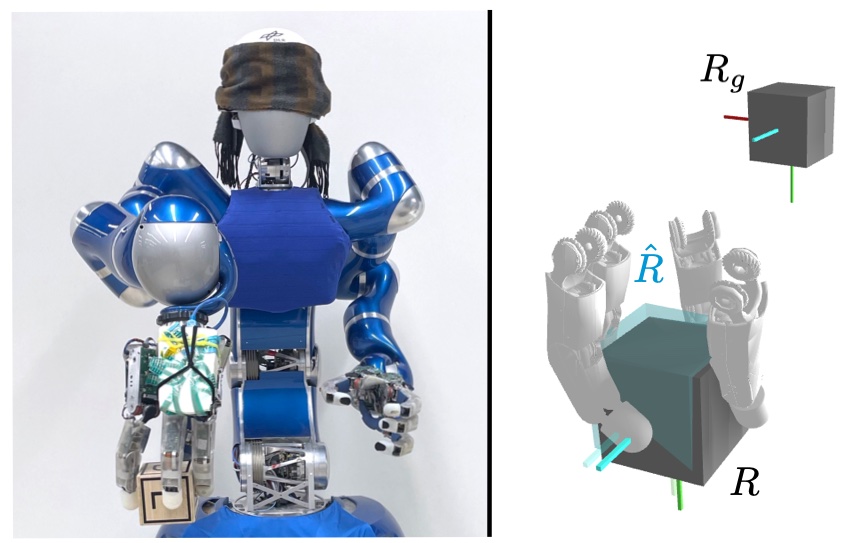}
  \caption{\textbf{Left:} Agile Justin~\cite{Bauml2014} performing in-hand manipulation while being blindfolded. 
  \textbf{Right:} The goal of the task is to bring the object orientation $R$ to a goal orientation $R_g$. 
  The estimated state $\hat{R}$ is visualized in transparent blue.\\}
  \label{fig:justin}
\end{figure}

However, for some problems, this approach is fundamentally insufficient for solving the task. 
In this paper, we identify blind, goal-conditioned in-hand manipulation (cf. \cref{fig:justin}) as one such problem; when no global information (such as from camera sensors) is available for tracking the manipulated object, many control sequences lead to stochastic outcomes and potentially unrecoverable loss of observability. While these control strategies are valid in settings where perfect state information is obtainable (like in a simulator), for deployment to realistic environments, a principled approach is required for obtaining controllers that are robust to such imperfect state information.

\begin{figure}
  \centering
  \includegraphics[width=.9\linewidth]{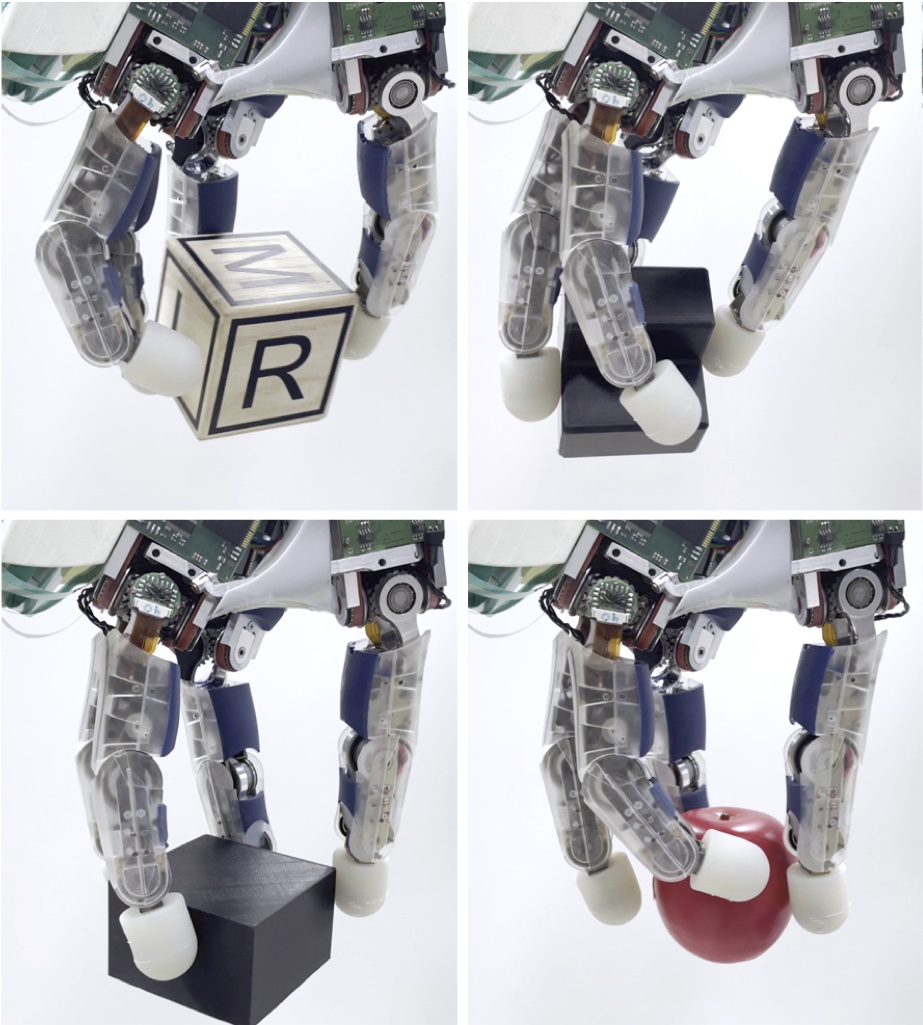}
  \caption{In-hand manipulation of different object shapes with the DLR-Hand II~\cite{Butterfass2001}. The object set and the benchmark results are depicted in \cref{fig:bench}. The task consists of deliberately reorienting the objects to an externally specified target orientation. The shown setting is especially challenging as the hand is oriented downwards, hence demanding permanent force closure. The manipulation is performed blindly, i.e., without cameras, using high-fidelity joint torque sensing for purely tactile tracking of the object pose.}
  \label{fig:title}
\end{figure}

\pagebreak

\subsection{Contributions}
\label{sec:contributions}
\begin{itemize}
    \item We identify pitfalls that arise when combining (learned) state estimators and controllers in the setting of purely tactile in-hand manipulation. 
    \item To solve these problems, we propose a unified training scheme for learning robust controllers and state estimators: by coupling the input of the policy to the predicted state during training, the policy becomes robust to estimation biases and is incentivized to avoid actions with unpredictable outcomes.
    \item In our experiments in simulation and in the real world, we show that the approach enables more extended, robust manipulation, including manipulation of various object shapes that were not shown in prior works in the setting of purely tactile goal-oriented in-hand manipulation (cf. \cref{fig:title}).
\end{itemize}

\subsection{Related Work}
\label{sec:related_workd}
Dextrous in-hand manipulation is a challenging control problem due to the complex nature of contact dynamics and the potentially high degrees of freedom encountered in these problems. Learning-based approaches and in particular reinforcement learning, have shown great potential for deriving in-hand manipulation policies~\cite{openai2018learning, Openai2019rubiks, Sievers2022, Qi2023hand, TouchDexterity, Khandate2023sampling, Pitz2023dextrous, Handa2023dextreme}. 
For example,~\citet{Sievers2022, Qi2023hand, TouchDexterity, Khandate2023sampling} learn controllers to rotate objects around fixed axes, where the policies are trained in simulation with domain randomization and transferred to the real system for deployment.

Critically, the approaches in these works are not goal-oriented, which is an essential requirement for many practical in-hand manipulation tasks.
Goal-oriented in-hand reorientation requires estimating the pose of the manipulated object to determine when the desired state is reached. 

For that, one line of research employs visual pose estimation based on one or multiple cameras~\cite{openai2018learning, Openai2019rubiks, Morgan2022complex, Handa2023dextreme}.
Visual input provides global information, which potentially enables unambiguously determining the object's pose. 
In~\citet{openai2018learning, Openai2019rubiks} and~\citet{Handa2023dextreme}, vision-based pose estimators are trained separately from the policy in simulation using extensive data augmentation and randomization of visual appearances. Consequently, to cover the possible state space of diverse lighting conditions, visual appearances, and camera angles encountered in practice, relatively large computational budgets are required.

Inspired by the human ability to blindfold and effortlessly reorient known objects inside its hand, we strive towards purely tactile in-hand manipulation, circumventing the abovementioned requirement for visual object tracking.

Our previous work on purely tactile, goal-conditioned in-hand manipulation combined a learning-based state estimator~\cite{Rostel2022learning} with a control policy trained in simulation in a modular architecture~\cite{Pitz2023dextrous}. However, because the estimator and policy were only combined towards the end of the training process, multiple stages of iterative refinement and finetuning, including multiple reward functions, were needed on the policy and estimator, respectively. 
Contrary to this, the unified approach proposed in this paper is more straightforward because it does not involve multiple training stages. Importantly, we implement our method as end-to-end hardware-accelerated, enabling orders of magnitude less training time.

\section{Estimator-Coupled Reinforcement Learning}
\label{sec:method}
\subsection{Motivation}
\label{sec:method_motivation}

\begin{figure*}
  \centering
  \includegraphics[width=\textwidth]{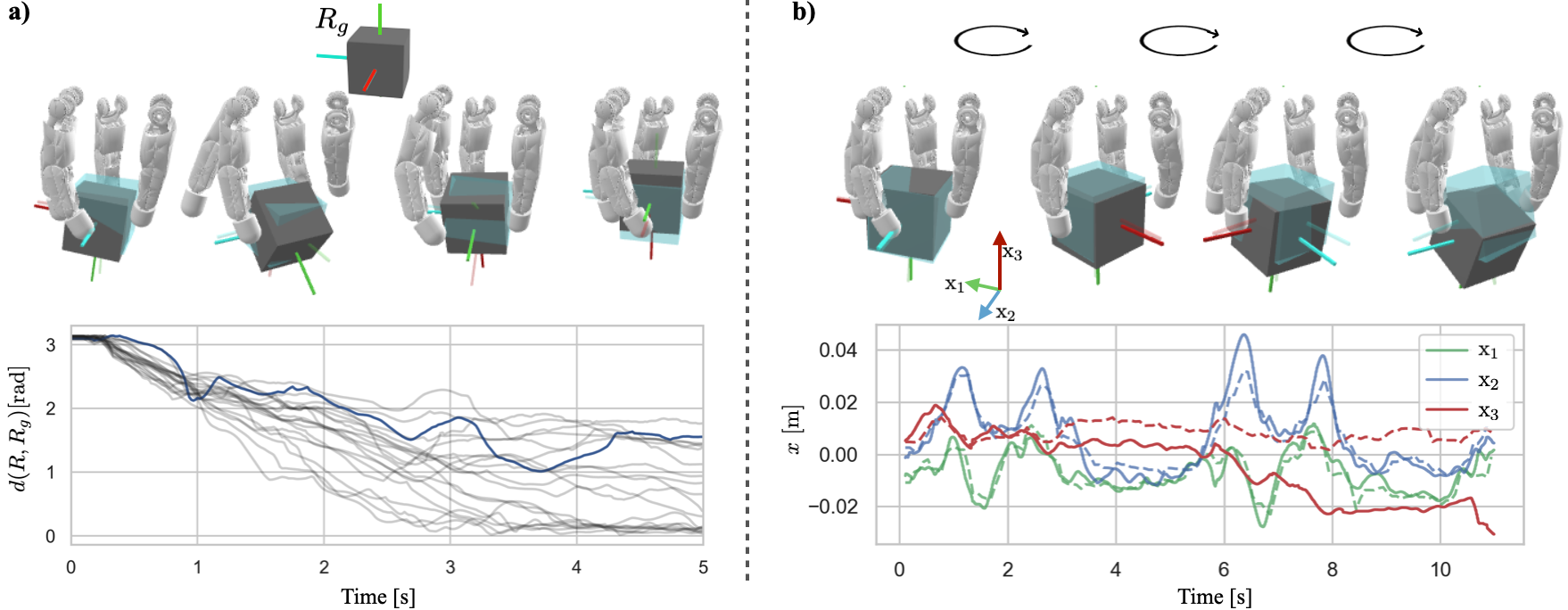}
  \caption{Two failure cases produced as rollouts of a non-robust control policy performing purely tactile in-hand reorientation of a cube. 
  \textbf{a)}: during reorientation towards a goal, the cube is held using only two fingers, allowing the cube to tip about the blue axis. Because the occurrence and extent of the tipping behavior depend, i.a., on unknowns, like friction effects, this effectively leads to \textit{stochastic dynamics} and consequently an ambiguous estimate of the object rotation. This is seen in the bottom figure, showing the angle to the target orientation $d(R,R_g)$ for multiple trials of the same rotation sequence in gray. The trajectory corresponding to the manipulation sequence shown on top is indicated in blue. Due to the object symmetry, these ambiguities in the estimate, in many cases, can not be resolved, leading to a failure of the task due to permanent loss of observability. 
  \textbf{b)}: the controller is tasked to perform rotations around the vertical axis (by setting new goals that are $\pi$\,rad rotated every 5s). Over the course of this manipulation, the estimated position (dashed line) drifts away from the ground truth position (solid line) in the $\mathrm{x_3}$-axis as it can hardly be determined by lateral contact measurements (compare \cite{Pitz2023dextrous}). The non-robust controller assumes the position to be unbiased, which leads to the cube being dropped.}
  \label{fig:motivational_failures}
\end{figure*}

The separation of concerns between state estimators and controllers is a well-proven design concept in robotic systems.
State estimators can aggregate past information over time, which in turn enables more concise control laws that only depend on the aggregated state.
In many partially observable domains, however, some parts of the state space may not be estimated unambiguously.
For example, in the setting of purely tactile manipulation, the pose of the manipulated object can only be estimated indirectly through contact sensing. 
Specifically, we identify two phenomena that frequently occur due to this lack of observability:
\begin{enumerate}
    \item Stochastic dynamics effects (\cref{fig:motivational_failures} a)) and
    \item Systematic biases/drifts in the estimate (\cref{fig:motivational_failures} b)).
\end{enumerate}

Control strategies that assume unbiased state estimates or produce unrecoverable ambiguities are, in general, not sufficient, which we discuss in \cref{sec:discussion}.
Instead, a robust control policy should adapt to both of these issues: it should 1. not rely on state inputs that are inherently unreliable and 2. avoid actions that cause stochastic dynamics with potentially permanent loss of state information. 

However, obtaining such robust control laws in practice is challenging, especially for the complex control task of dextrous in-hand manipulation.
Expanding upon recent work on learning-based controllers for tactile in-hand manipulation, in this paper we develop a method for learning such robust control strategies.

We find that the training of state estimators and controllers from data gives rise to a chicken-and-egg-problem:
Data for training of a state estimator can only be obtained by unrolling a suitable controller. On the other hand, learning a controller accounting for 1. and avoiding 2. needs access to a state estimator. 
This results in a coupled optimization problem, where one component cannot be trained without access to the other.

In this paper, we address this problem by a coupled learning approach, Estimator-Coupled Reinforcement Learning (EcRL), where we concurrently train a policy and a state estimator from data collected in simulation. The policy, conditioned on the estimator output, is trained by reinforcement learning to maximize the task reward while the estimator is trained supervised to predict task-relevant quantities.

In the remainder of this section, we describe in more detail the proposed unified optimization scheme (\cref{sec:concurrent_learning}) as well as the training of the individual components: the state estimator (\cref{sec:estimator_learning}) and the reinforcement learning (\cref{sec:policy_learning}).

\subsection{Concurrent Learning Scheme}
\label{sec:concurrent_learning}
We concurrently train two components: a \textit{state estimator} which, given an observable state $z_t$ and control input $u_t$ at each timestep, produces an estimate of the task-relevant system state $\hat{s}_t$
\begin{equation}
    f_{\phi}(z_t, u_{t-1}, \hat{s}_{t-1}) = \hat{s}_{t}
\end{equation}
parametrized by learnable parameters $\phi$,
and a stochastic, goal-conditioned control policy

\begin{equation}
    u_t \sim \pi_{\varphi}(\cdot| z_t, \hat{s}_t, s_g)
\end{equation}
conditioned on goal-state $s_g$ with learnable parameters $\varphi$. 

\begin{figure}
\centering
\includegraphics[width=70mm]{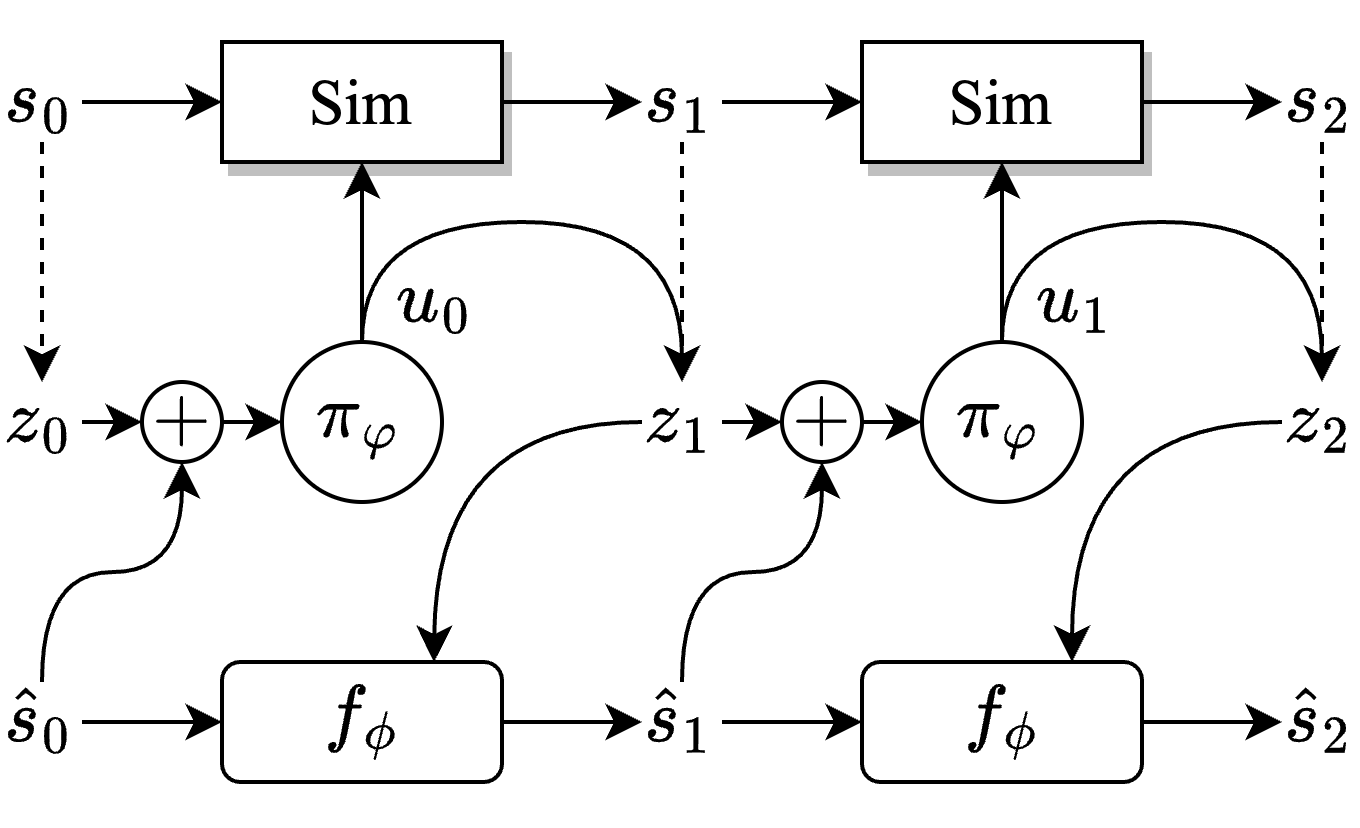}
\caption{The system state $s$ is advanced in time by the simulator or the real robot. From the observation $z$, the state estimator $f_{\phi}$ recurrently produces an estimate of the state $\hat{s}$. Based in the current observation and the predicted state, the policy $\pi_{\varphi}$ computes control inputs $u$.}
\label{fig:stepping}
\end{figure}

Starting from randomly initialized $f_{\phi}$ and $\pi_{\varphi}$, we alternately collect training data $\mathcal{D}$ by performing rollouts of length $T$ using $f_{\phi}$ and $\pi_{\varphi}$ (see \cref{fig:stepping}) and perform gradient-based optimization on the collected data. The policy is trained by on-policy, model-free reinforcement learning on sequences of tuples $(r_t, (z, \hat{s}, s_g)_t, (z, \hat{s}, s_g)_{t+1}, u_t)$ from the collected rollout. The estimator is trained supervised to regress the task state $s$, which is available from the simulator during training.

Because inputs received from a randomly initialized estimator may induce a value assignment problem for the policy in the early learning stages, we optionally give the ground truth state $s_t$ as input to the policy during training with a probability $\rho \in [0,1]$ for a given rollout sequence. In our experiments, we decrease the probability from $\rho=1.0$ to $\rho=0$ in the early training phase using a linear schedule.
The algorithm is summarized in \cref{algo:training}, where we omitted the dependence on goal state $s_g$ for brevity.
\subsection{Estimator Learning}
\label{sec:estimator_learning}
Given the history of observations $z_{0:t}$ and an initial state $s_0$, the estimator gives a prediction of the current system state $s_t$.
Here, we use a simple recursive model of the form

\begin{equation}
    \begin{bmatrix}\hat{s}_t\\l_t\end{bmatrix} = f_{\phi}\left(z_t, u_{t-1}, \begin{bmatrix}\hat{s}_{t-1}\\l_{t-1}\end{bmatrix}\right) \boxplus
    \begin{bmatrix}\hat{s}_{t-1}\\l_{t-1}\end{bmatrix}
\end{equation}

where $\boxplus$ is the generalized addition operator \cite{hertzberg2013integrating} (see also \cite{Rostel2022learning}), and $l_t \in \mathrm{R}^{d_l}$ is a $d_l$-dimensional latent vector. The intention of augmenting the state space by latent dimensions is to give the estimator the ability to model effects that are not directly part of the regressed state $s$. The advantage of this parametrization over a recurrent neural network is that the estimator can be conveniently initialized to a known system state $s_0$ during training and deployment.

The state estimator is then optimized by backpropagation through time on sequences drawn from $\mathcal{D}$, minimizing the root mean squared error (rmse) between the ground truth state $s$ and the predicted state $\hat{s}$.

In practice, we divide the sequences stored in $\mathcal{D}$ into $b$ minibatches of size $|\mathcal{D}|/b$ and perform $k * b$ gradient steps, where $k$ is a hyperparameter controlling data reusage.

While more involved approaches, including Differential Bayesian Filters \cite{Rostel2022learning}, can be used to explicitly model uncertainties, this paper focuses on investigating the interactions of the state estimator with the controller rather than the specific estimator architecture. 

\subsection{Policy Learning}
\label{sec:policy_learning}
The control policy $\pi$ is obtained by reinforcement learning.
We search for a policy $\pi_{\varphi}(u_t| z_t, \hat{s}_t, s_g)$ which maximizes the expected discounted sum of rewards $\mathbb{E} \left[\sum_t \gamma^t r_t\right]$, where the task is encoded in a reward function $r_t = r(s_t, s_g)$. 
We describe the learning setting specific to tactile in-hand manipulation in \cref{sec:rl}.

Notice that, if predictions from the most recent estimator model are used as input to the policy, the learning problem is non-stationary.
As samples collected from previous estimator iterations become less informative, the problem lends itself towards on-policy optimization.

Here we choose Proximal Policy Optimization (PPO) \cite{Schulman2017proximal} as a stable and performant on-policy algorithm for learning the policy $\pi$ and a value function $V(z_t, \hat{s}_t, s_g)$.

We summarize the hyperparameters used in \cref{tab:hyper}. Notably, we use an adaptive learning rate schedule \cite{Petrenko2023dexpbt} and a feedforward policy with a dense architecture \cite{Sinha2020d2rl}.

\begin{algorithm}
  \footnotesize
 \caption{Estimator-coupled learning of policy and estimator EcRL}
 \begin{algorithmic}[1]
\State {Initialize policy $\pi$ and estimator $f$}
\State {Initialize state $\hat{s}_0$}
  \While{not converged}
    \State {Initialize Buffer $\mathcal{D} = \emptyset$}
    \For{each environment}
    \State $k \sim U(0,1)$
    \For{$ t = 0, 1, 2, ..., T$}
      \If{$k < \rho$ \textbf{or} $t=0$}
        \State $\hat{s}_{t} = s_{t}$ \algorithmiccomment{\textit{use ground truth state}}
      \Else
        \State $\hat{s}_{t} = f(z_{t}, u_{t-1}, \hat{s}_{t-1})$ \algorithmiccomment{\textit{use estimate}}
      \EndIf
      \State $u_{t} \sim \pi(z_{t}, \hat{s}_{t})$
      \State $r_{t}, z_{t+1}, s_{t+1} \leftarrow \text{Sim}(u_{t}) $  \algorithmiccomment{step simulator}
      \If{reset simulator}
        \State $\hat{s}_t = \hat{s}_0$
      \EndIf
      \EndFor
      \State $\mathcal{D}  \leftarrow \mathcal{D} \cup \{(r_{t}, u_{t}, z_{t+1}, s_{t+1})_{t=0}^T\}$
    \EndFor
    \State Update $f_{\phi}$ on data from $\mathcal{D}$
    \State Update $\pi_{\varphi}$ on data from $\mathcal{D}$
    \State $\rho \leftarrow \max (\rho - \delta_{\rho}, 0)$ \algorithmiccomment{update schedule}
  \EndWhile
 \end{algorithmic}
 \label{algo:training}
 \end{algorithm}

\section{Application to blind, goal-conditioned in-hand manipulation}
\label{sec:application}

Achieving robust dextrous in-hand manipulation with multi-fingered hands is a significant control challenge, especially in the absence of visual input. Goal-conditioned reorientation requires particularly deliberate control strategies, as even slight control errors can lead to imminent task failure, as shown in the motivation examples \cref{fig:motivational_failures}.

In this section, we describe the application of the training scheme described in \cref{sec:method} for obtaining robust controllers in this setting. We begin by outlining the studied task of in-hand reorientation in \cref{sec:task}. We then describe the system and hardware used in our experiments in \cref{sec:system}, the reinforcement task design (\cref{sec:rl}), and the simulation setup in which the training is conducted (\cref{sec:sim}).

\subsection{Task Description}
\label{sec:task}

Given an initial object orientation $R_0$, the goal of the control policy is to reorient an object to an externally specified target orientation $R_g$. 

As in prior work \cite{Pitz2023dextrous}, we constrain the possible goal orientations to be on a $\pi/2$-discretized subspace of $\mathrm{SO}(3)$ (octahedral group), resulting in 24 possible goal orientations. This discretization uniformly covers the space of rotations, allowing for systematically analyzing the performance on individual target orientations during training and benchmarking (see \cref{fig:bench}).

We define the estimated system state $\hat{s}$ to consist of object position $\hat{x}$, object rotation $\hat{R}$ and the corresponding velocities $\hat{v}$ and $\hat{w}$. 

\subsection{System Description}
\label{sec:system}

Building on our previous work~\cite{Sievers2022, Pitz2023dextrous}, we use the torque-controlled DLR-Hand~II~\cite{Butterfass2001} as our robotic hardware.
The hand has four identical fingers, each with three actuated joints and a fourth and last joint that is coupled to the previous one, resulting in a total of $N_{\mathrm{dof}}=12$ degrees of freedom.
The outputs of the control policy are interpreted as desired joint angles $q_d$, queried at a rate of $10$\,Hz. The target joint angles are low-pass filtered and given to an underlying impedance controller at $1000$\,Hz. Using this impedance control, contacts can be inferred from measuring the difference between the current measured joint position $q$ and the desired joint angles $q_d$, making use of the high-fidelity joint torque sensing of the DLR-Hand~II. No further tactile information is used. We refer to~\citet{Sievers2022} for a detailed description of the control architecture.

\subsection{Reinforcement Learning Environment}
\label{sec:rl}

The reinforcement learning setup directly follows from the in-hand reorientation task described in \cref{sec:task} with some modifications to speed up training and ease the sim-to-real transfer. We will outline the specific setting used in the following.

\subsubsection{Episode}
At the beginning of each learning episode, we initialize the state of the hand-object system with a stable grasp.
To cope with a variety of considered object geometries, in practice, we precompute a set of grasping configurations by executing a simple heuristic. The heuristic consists of 1) spawning the object inside the opened hand with gravity disabled, 2) commanding target actions to close the fingers, 3) enabling gravity, and 4) checking if the change in object pose is below a set threshold. The advantage of this approach is that it is object-agnostic and can easily be replicated on the real system using the inbuilt impedance controller.

In contrast to~\citet{Khandate2023sampling}, we found it sufficient to work with a small set of stable grasps around a single initial pose. Also, we found it not necessary to use a gravity schedule in our experiments.

During training, a new target orientation is randomly sampled at a fixed interval of $T_g=5$\,s. At the end of this interval, the target orientation is considered reached if the angle between  $R_{g}$ and the actual rotation $R$ is below a threshold of $0.4$\,rad. In the case of success, a new goal is sampled, and the episode continues. We find that this incentivizes the policy to end in stable and well-controlled grasps.

The environment is terminated if
\begin{itemize}
    \item the distance of the object from the origin is greater than $0.1$m or
    \item the goal is not reached after $T_g$ or
    \item a set timeout of $T_{max} = 20s$ is reached.
\end{itemize}
Additionally, during the learning phase only, we terminate the episode if the angle between the estimated orientation and the ground truth orientation exceeds a set threshold of $\pi/4$, which discourages the policy from causing unrecoverable loss of observability.

\subsubsection{Observations}
\label{sec:observations}
The observed state $z_t$ given to the policy and the estimator consists of measured joint angle $q \in \mathbb{R}^{N_\mathrm{dof}}$ and control error $e_q = q - q_d$. 

To increase the temporal resolution, we stack 6 of these measurements at a rate of 60\,Hz, yielding a history of 0.1\,s per observation vector. The estimator and control policy are then executed at a frequency of 10\,Hz.

For the policy, we additionally give the estimated object position~$\hat{x}$ and the estimated relative rotation to the goal~$\hat{R}_\Delta = R_\text{g}^{-1}\hat{R}$ for goal-conditioning.

\subsubsection{Reward}
The agent is rewarded based on the difference between the angle to the goal orientation in the current timestep $\theta_{t}$ and the previous timestep $\theta_{t-1}$ and a positive reward is given whenever the $\theta_{t} < \theta_{t-1}$. 
Furthermore, we penalize deviations of the current position of the object~$x_t$ from the mean initial position~$\bar{x}_0$ and deviations of the current joint positions~$q_t$ from the mean initial joint position~$\bar{q}_0$.

In summary, the reward function is
\begin{align}
  r_t = \quad\; &\lambda_{\theta} \min \left( \theta_{t-1} - \theta_{t}, \theta_\text{clip} \right) \nonumber \\
             - & \lambda_{x'} \left( \lambda_{x} \min \left( \| x_t - \bar{x}_0 \|, x_\text{clip} \right)  \right)^4 \\
             - &\lambda_{q}\frac{1}{N_\mathrm{dof}} \| ( q_t - \bar{q}_0 )^4 \|_1 \nonumber 
\end{align}
with coefficients $\lambda_{\theta}= 1000$, $\lambda_{x'} = 0.1$, $\lambda_{x}= 50$, $\lambda_{q} = 2000$, $\theta_\text{clip} = 0.1$\,rad, $x_\text{clip} = 0.3$\,m. 
Limiting the maximally available reward for rotations is a simple but significant modification that forces the agent to rotate the object slowly and makes the resulting policies more suitable for a robust Sim2Real transfer. The position clipping was introduced to mitigate the effects of potential simulation instabilities on the policy gradient.

\subsection{Simulation Environment}
\label{sec:sim}

Training is conducted using the IsaacSim GPU-accelerated simulator~\cite{IsaacSim}. This allows for parallelization of the loop over environments in \cref{algo:training}, where we use 4096 parallel environment instances with unique sets of domain randomization parameters in our experiments.

The general setup of the DLR-Hand~II inside the simulator is modeled analogously to our previous work \cite{Sievers2022}. However, in order to achieve a successful Sim2Real, we put significant effort into ensuring that contacts are faithfully simulated in IsaacSim, including accurate modeling of colliding geometries and significant randomization of friction parameters. 

Other aspects of the domain randomization are similar to the parametrization described in ~\citet{Pitz2023dextrous}, including observation noise, measurement biases, randomization of object properties like size and mass as well as randomization of control parameters.

Additionally, before deploying the learned controllers to the real system, we finetune in a setting where sporadic, randomly directed forces and torques are applied to the object, as is a common practice for accounting for any unmodeled effects in reality \cite{Openai2019rubiks}.

Training from scratch until convergence per object takes 6.5h (median) on a single NVIDIA T4 GPU, where, on average, 95\% of final performance is reached after 5h when the ground truth state schedule finishes.

\begin{table}
    \caption{Hyperparameters}
    \label{tab:hyper}
    \centering
    \begin{tabular}{c c}
      \toprule
         Hyperparameter & Value \\
         \midrule
        \multicolumn{2}{c}{EcRL training parameters}\\
        \midrule
        $\rho_0 $ &  $1$\\
        $\delta_{\rho} $ &  $\num{1e-3}$\\
        rollout length $T$ & 32 \\
        \midrule
        \multicolumn{2}{c}{Reinforcement learning paramaters}\\
        \midrule
        learning rate & adaptive (\cite{Petrenko2023dexpbt})  \\
        hidden layers & [512, 512, 256, 128] \\
        minibatch size & $2^{15}$\\
        $\epsilon_{clip}$ \cite{Schulman2017proximal} & 0.2 \\
        entropy coef \cite{Schulman2017proximal}& $\num{1e-3}$\\
        $\tau$ & $0.95$\\
        $\gamma$ & $0.99$ \\
        \midrule
        \multicolumn{2}{c}{Estimator parameters}\\
        \midrule
        learning rate & $\num{5e-4}$ \\
        hidden layers & [512, 512, 512, 512] \\
        minibatch size & $2^{10}$ \\
        latent dims $d_l$ & 32 \\
        data reuse $k$ & 2\\
      \bottomrule
    \end{tabular}
\end{table}

\section{Results}
\label{sec:results}
We now present the results of applying the proposed estimator-coupled reinforcement learning approach to the task of purely tactile, goal-conditioned in-hand manipulation. In the following, we first introduce the considered baselines (\cref{sec:baselines}) and the methods of evaluation (\cref{sec:eval_sim}).

\subsection{Baselines}
\label{sec:baselines}
Using the same general setting described above, we compare the following training methods:
\begin{itemize}
    \item \textbf{Naive}: We train an ``oracle'' policy on ground truth state input until convergence, corresponding to the case $\rho = 1$, $\delta_{\rho} = 0$ in \cref{algo:training}. The estimator is continually trained on the rollouts of the policy. Only at test time the output of the estimator is given as input to the policy. 
    \item \textbf{EstimAda}: We freeze the oracle policy obtained in the previous step but iteratively train an estimator, where the estimated state is given back as input to the policy used for generating the data. This approach is akin to rapid motor adaption approaches (see, e.g., \citet{Qi2023hand}) and more directly comparable to the iterative approach in \citet{Pitz2023dextrous}.
    \item \textbf{EcRL} (ours): We train policy and estimator concurrently using the coupled learning approach described in \cref{sec:method}.
\end{itemize}
We give an overview of the hyperparameters used for all experiments in \cref{tab:hyper}.  

\subsection{Evaluation in Simulation}
\label{sec:eval_sim}
We evaluate the performance of the learned policies by performing a  benchmark task covering all discrete goal orientations, in line with prior work \cite{Pitz2023dextrous}. We perform 50 trials for each of the 24 goal orientations, resulting in a total of 1200 rollouts, for each of which domain randomization parameters are sampled from the same distribution as in training but with a different random seed. We then report the percentage of goals reached within time limit $T_g$ as success rate $B$.

We consider a set of 4 different objects (see right column in \cref{fig:bench}):
\begin{itemize}
    \item cube with edge length $8$\,cm, as used in previous works \citet{Sievers2022, Pitz2023dextrous}
    \item cuboid with extends 8x8x5\,cm 
    \item non-convex, ``L''-shaped object
    \item plastic apple as part of the YCB dataset \cite{YCB}.
\end{itemize}
For each of the objects, we train separate policies using the approaches described in \cref{sec:baselines}.

\subsection{Quantitative Comparison}

\begin{figure}
  \centering
  \includegraphics[width=\linewidth]{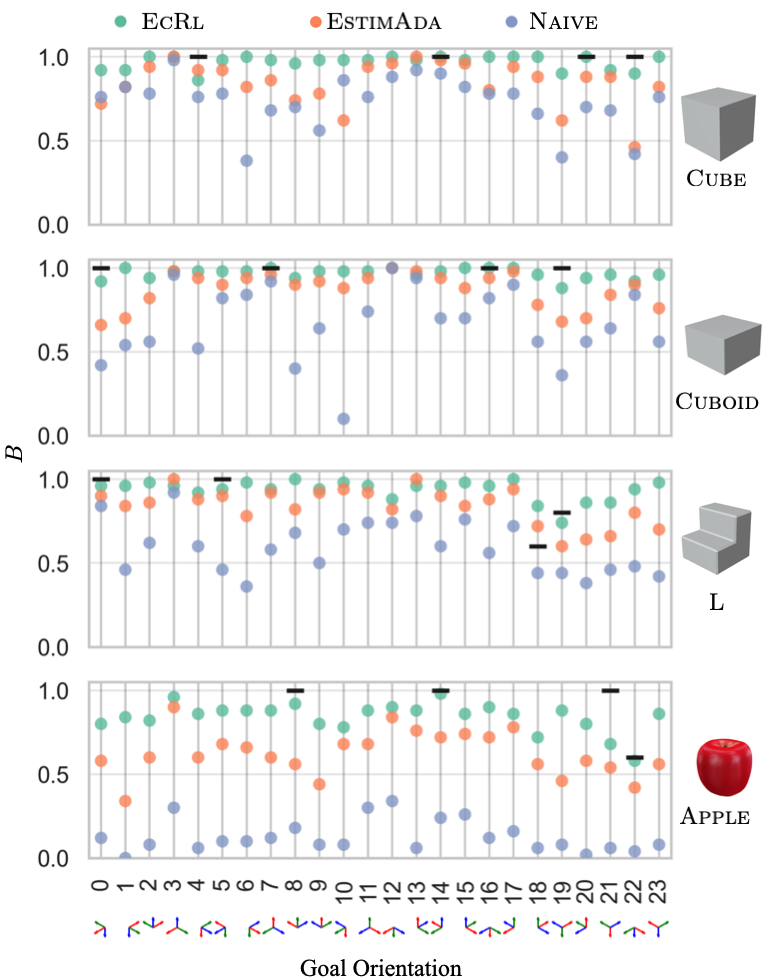}
  \caption{Success rates $B$ on the in-hand reorientation benchmark for each of the 24 possible goal rotations column-wise and each of the considered objects row-wise. 
  The goal indices, as well as the associated coordinate transformation, are indicated on the horizontal axis. 
  Note that index 3 is the identity rotation for which the controller needs to hold the object for 5 seconds without rotating it. 
  The four considered object shapes are shown on the right. 
  Each dot represents the success rate over 50 trials in simulation. 
  Black horizontal bars indicate results of \textsc{EcRL} on the real system for the 2 goal orientations which perform best and worst respectively in simulation for each object.}
  \label{fig:bench}
\end{figure}

We compare the final performances of the considered methods on the benchmark task in simulation in \cref{tab:eval_bench}. Note that the \textit{Oracle} baseline receives the privileged ground truth state input also at test time. It can be seen that the learning scheme proposed in this paper results in estimators with lower prediction error and policies that consistently exceed the performance of those of the baselines.
In \cref{fig:bench}, we also show a breakdown of the performances on the individual goal rotations.

\begin{figure}
  \centering
  \includegraphics[width=0.8\linewidth]{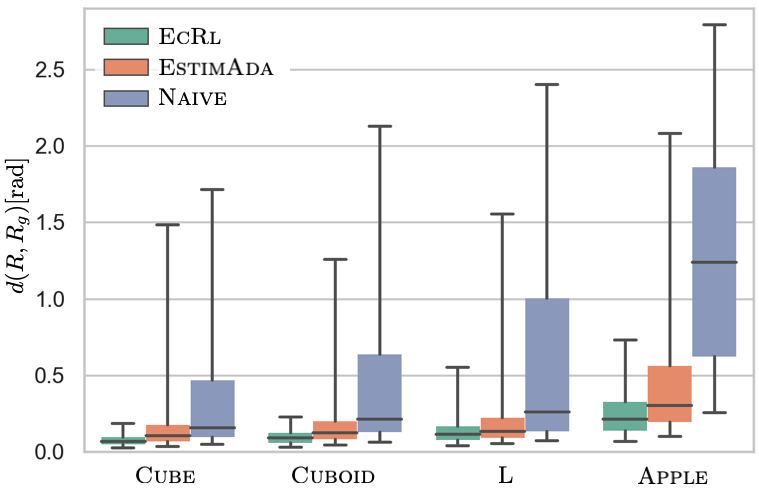}
  \caption{Distribution of angles $d(R, R_g)$ between the actual object orientation $R$ and goal orientation $R_g$ as measured in the final timestep $T_g$ in the benchmark setting. Boxes cover the first and third quartiles, and whiskers represent the 5th and 95th percentiles.}
  \label{fig:angle_plot}
\end{figure}

Of the objects studied, the performance in terms of success rate is lowest for the apple. This can be explained by the rotational symmetries that make estimating absolute orientation (which is necessary for accomplishing the task) difficult for any tactile state estimator due to rotational drift. Still, especially in the case of the apple, we observe that our estimator-aware training procedure leads to more deliberate manipulation behavior, resulting in improvements in predictability and overall higher success rates. 
This effect is also observed in the distribution of final angles to the goal orientation in \cref{fig:angle_plot}, indicating that \textsc{EcRL} produces more reliable reorientation results.

While the benchmark task measures the capability of performing a single reorientation from an initial to a goal orientation, for the cube, we find that it is even possible to perform multiple consecutive reorientations without intermediate visual grounding. A quantitative evaluation of this is shown in \cref{tab:eval_consec}. The results of the \textsc{EcRL} method show that extended manipulation sequences are possible with the learned robust control strategies, even without visual feedback. 

\begin{table}
    \centering
    \caption{Evaluation on Reorientation Benchmark}
    \begin{tabular}{l l c c c}
      \toprule
         \quad &\quad & \makecell{Success Rate\\ $B$ [\%]} & \makecell{Estimator Prediction\\ Error [rad]}\\
         \midrule
        Cube &\textcolor{gray}{Oracle}  & \textcolor{gray}{99}  &  \textcolor{gray}{\tentry{0.09}{0.07}} \\
        &Naive  & 73 &  \tentry{0.24}{0.30}\\
        &EstimAda  & 84  & \tentry{0.19}{0.22} \\
        &EcRL (ours) & 97  & \tentry{0.12}{0.13}\\
        \midrule
        Cuboid &\textcolor{gray}{Oracle}  & \textcolor{gray}{99}  &  \textcolor{gray}{\tentry{0.11}{0.10}} \\
        &Naive  &  67 &   \tentry{0.32}{0.40}\\
        &EstimAda  & 87  & \tentry{0.18}{0.18} \\
        &EcRL (ours) & 97  &  \tentry{0.12}{0.10}\\
        \midrule
        L &\textcolor{gray}{Oracle}  & \textcolor{gray}{98}  &  \textcolor{gray}{\tentry{0.16}{0.16}} \\
        &Naive  &  59 &   \tentry{0.32}{0.39}\\
        &EstimAda  & 84&  \tentry{0.18}{0.17} \\
        &EcRL (ours) & 94  &   \tentry{0.13}{0.13}\\
        \midrule
        Apple &\textcolor{gray}{Oracle}  & \textcolor{gray}{99}  &  \textcolor{gray}{\tentry{0.17}{0.14}} \\
        &Naive  &  13&   \tentry{0.51}{0.44}\\
        &EstimAda  & 62&  \tentry{0.29}{0.21} \\
        &EcRL (ours) & 84&   \tentry{0.22}{0.16}\\
        
      \bottomrule
    \end{tabular}
    \label{tab:eval_bench}
\end{table}

\begin{table}
    \centering
    \caption{Cube Consecutive Successes}
    \begin{tabular}{l | c c c c | c}

        \quad & \textcolor{gray}{Oracle} & Naive & EstimAda & \makecell{EcRL \\ (sim)} &\makecell{ EcRL \\(real)} \\
        \midrule
        \makecell{\# Successes\\ (median)}  &  \textcolor{gray}{ $> 100$ \tablefootnote{The actual number of median consecutive successes produced by the oracle policy is $>100$,  we stopped the simulation after that point.}}  & 1 & 2 & 10 & \makecell{9  \\ \makecell{\tiny{[14, 12, 30, 3, 4} \\ \tiny{5, 3, 16, 20, 6]}}}\\

    \end{tabular}
    \label{tab:eval_consec}
\end{table}

\subsection{Real World Experiments}
\label{sec:eval_real}
\begin{figure*}
  \centering

  \includegraphics[width=0.9\textwidth]{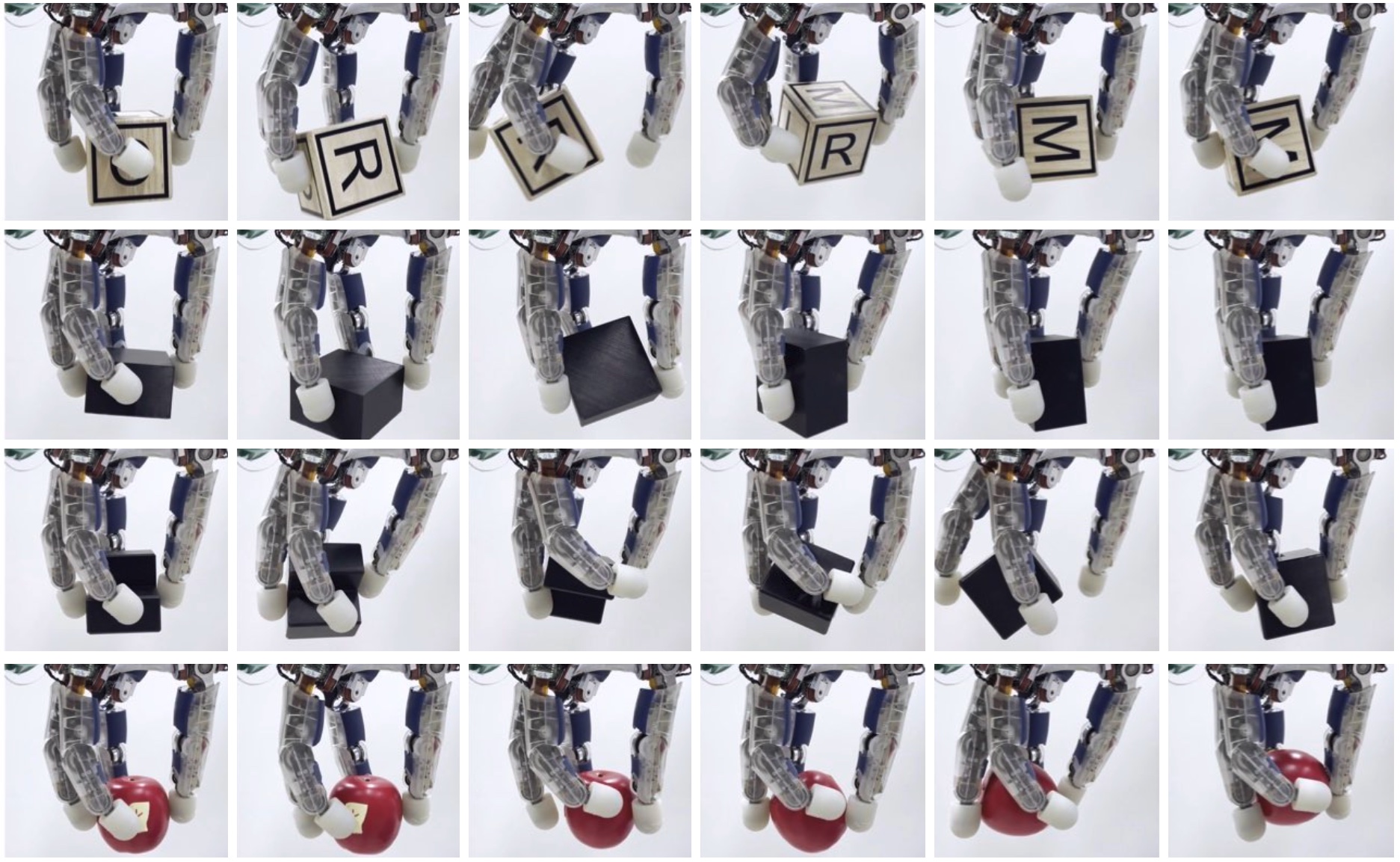}

  \caption{Real-world rollouts for all four objects showing successful reorientation to a given target orientation (goal index 20, compare \cref{fig:bench}). Full videos can be found on the project website \href{\gitlink}{\texttt{\gitlink}}.}
  \label{fig:real_strip}
\end{figure*}

To evaluate the transferability of the policies learned by \textsc{EcRL} to the real system, we evaluate a subset of the reorientation benchmark in \cref{fig:bench}. 
Note that, for the goal orientations which suggest strong performance in simulation, we also observe $100\%$ success rates on the real system.
In \cref{fig:real_strip}, we show a manipulation sequence for each of the 4 objects, reaching the desired goal orientation.
We also provide results for 10 trials with randomly choosen sequences of goals on the consecutive cube reorientation task in \cref*{tab:eval_consec}, showing that the learned policies transfer well to the real real system, enabling extended manipulation sequences.
For videos of real-world demonstrations, please visit the project website and examine the supplementary video. 

\section{Discussion}
\label{sec:discussion}
Comparison of the performance metrics in \cref{tab:eval_bench} and \cref{fig:bench} reveals a significant advantage of the proposed approach over the naive training approach. We hypothesize that this performance gap can be decomposed into two effects: 1) the \textit{estimator gap}, caused by model errors of the estimator, and 2) the \textit{policy gap}, caused by non-robust control strategies.

\subsection{Estimator Gap}
For a control policy conditioned on an estimated state, a low prediction quality clearly can lead to a loss in performance. Interestingly, we observe that the prediction error is significantly lower during the evaluation of the oracle policy with privileged input when compared to the other approaches (\cref{tab:eval_bench}). We hypothesize that the drop in prediction quality at test time (corresponding to the \textit{naive} approach) can be attributed to either a) biased estimates, where the estimator learned to infer states from the outputs of the oracle policy directly, or b) distribution shifts due to change in policy behavior induced by transitioning from the oracle state to the estimated state as input to the policy.

In principle, both effects can be mitigated by training the estimator on data generated by a policy receiving the estimated input, as in the \textsc{EstimAda} approach.

Consequently, assuming that the \textsc{EstimAda} training scheme fully closes the estimator gap, the performance difference remaining to reach the performance of the estimator-coupling approach can only be caused by shortcomings of the controller.

\subsection{Policy Gap}

We conjecture that the performance difference between \textsc{EstimAda} and \textsc{EcRL} quantitatively shows the effects of inherently non-robust control strategies as anecdotally described in \cref{fig:motivational_failures}.
As an example case for investigating the qualitative differences between the control strategies learned by EcRL and EstimAda, we closely inspected both policies when continuously setting new rotating goals around the $\mathrm{x_3}$ axis (compare \cref{fig:motivational_failures} b)). The \textsc{EcRL} policy only rarely lifts more than one finger at a time to perform the rotation, while the \textsc{EstimAda} policy, displayed in the sequence shown in \cref{fig:motivational_failures}, lifts two fingers.
Because it is difficult to show this effect on paper, we encourage the reader to examine the videos and interactive animations on the project website.
We believe that the \textit{policy gap} can only be resolved by accounting for the limitations of the state estimator during the policy training, which our coupled training framework fulfills. 

\section{Conclusions}

This work investigates the interplay between state estimators and controllers in the setting of purely tactile goal-conditioned object reorientation. To obtain robust controllers in this setting, we find that accounting for the limits of state estimation is essential. Consequently, we find that policies trained only from privileged input often lack robustness. 

To address these issues, we propose a unified learning scheme that allows the policy to receive the estimated state during training. Our experiments show that this empirically leads to higher end-to-end performance on the studied in-hand reorientation task.

We show that policies for manipulating various objects can be obtained within only a few hours of training due to our unified, highly efficient, parallelized implementation. Importantly, we show that the obtained controllers are robustly transferable to the real system. 
The demonstrated performance on the DLR Hand-II in this challenging task fundamentally exceeds all previous work with respect to the diversity of manipulated objects and robustness, with a median of nine consecutive reorientations for the cube.
In the future, we want to consider explicitly modeling uncertainties in the estimated state and reorient unknown objects with a single policy.

\scriptsize
\bibliographystyle{IEEEtranN-modified}
\bibliography{IEEEabrv, bibliography}

\end{document}